\documentclass{Interspeech2024}
\usepackage{pifont}
\usepackage{multirow}

\interspeechcameraready

\title{Evaluation of data inconsistency for multi-modal sentiment analysis}

\name{Yufei}{Wang}
\name{Mengyue}{Wu}

\address{
  $^1$Shanghai Jiao Tong University, China} 
\email{arthur-w@sjtu.edu.cn, mengyuewu@sjtu.edu.cn} 

\keywords{multi-modal sentiment analysis, multi-modal large language model, data inconsistency}

\begin{document}

\maketitle

\begin{abstract}
    Emotion semantic inconsistency is an ubiquitous challenge in multi-modal sentiment analysis (MSA). MSA involves analyzing sentiment expressed across various modalities like text, audio, and videos. Each modality may convey distinct aspects of sentiment, due to subtle and nuanced expression of human beings, leading to inconsistency, which may hinder the prediction of artificial agents. In this work, we introduce a modality conflicting test set and assess the performance of both traditional multi-modal sentiment analysis models and multi-modal large language models (MLLMs). Our findings reveal significant performance degradation across traditional models when confronted with semantically conflicting data and point out the drawbacks of MLLMs when handling multi-modal emotion analysis. Our research presents a new challenge and offer valuable insights for the future development of sentiment analysis systems.
    
\end{abstract}

\section{Introduction}

\begin{figure}[t]
  \centering
  \includegraphics[width=0.9\linewidth]{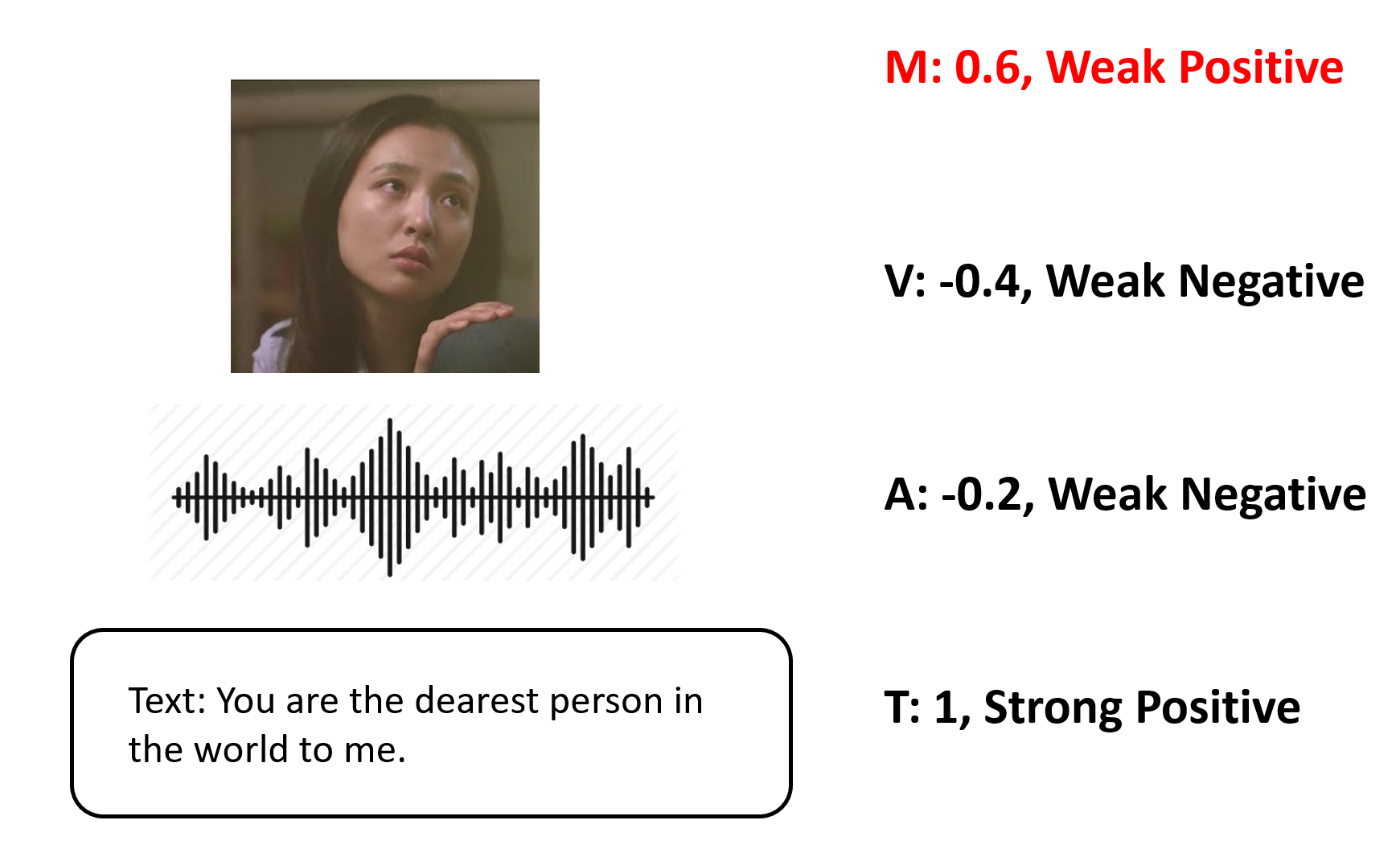}
  \caption{An example of multi-modal data conflicting samples. "M", "V", "A", "T" represent multimodal, visual, acoustic, textual label respectively.} 
  \label{fig:conflict}
\end{figure}

Multi-modal sentiment analysis has become a research focus in the field of artificial intelligence over the last few years in human-machine interaction~\cite{ kaur2022multimodal,yang2020cm}.
Its primary aim is to decipher the sentiment expressed by individuals across a spectrum of modalities, including language, facial expressions, speech, posture and sometimes physiological signals~\cite{li2022hybrid}. 
However, despite the rapid advancements in multimodal learning,
the task of detecting human emotions remains challenging, given the intricate nature of emotion expression~\cite{franceschini2022multimodal}, that
emotion expression is always subtle and nuanced.
This subtlety, which is complicated yet very common, comes from the intricate interplay of cultural, societal, and personal factors that shape how individuals perceive and convey their feelings.
Thus, human emotions expressed from different modalities tends to be inconsistent. An illustration of conflicting emotion sample is shown in Figure~\ref{fig:conflict}. 


In multi-modal sentiment analysis, the three most commonly utilized modalities are text, audio and visual~\cite{ghosal2018contextual}. 
Textual data plays a pivotal role, as it carries significant semantic context embedded in language~\cite{wu2021text}. The work~\cite{huang2021text} combines BERT~\cite{kenton2019bert} and CNN and acquires remarkable performance on movie review emotion classification dataset. 
However, depending solely on textual data may not suffice, as individuals also rely on non-verbal expressions to convey emotions, like acoustic and visual. 
By incorporating audio and visual cues, models can better capture the interplay between these modalities and have a broader recognition of the context. 
In Geometric Multimodal Contrastive (GMC) learning~\cite{poklukar2022geometric}, they compare the model performance using only uni-modal modality with multiple modalities and prove that multiple modalities can significantly improve the accuracy. 
The works~\cite{lv2021progressive, sun2023layer} introduce a message hub to exchange information with each modality to merge information from uni-modals and result a better performance compared with uni-modal classification. The recent benchmark~\cite{lian2024merbench} compares the uni-modal and multimodal emotion classification results for multiple datasets using different feature extractors. 
With the rise of large language models (LLMs), embedding other modalities into LLMs becoming popular.
Multimodal Large Language Models (MLLMs)
accommodate the audio-video-text multi-modal data together with LLMs ~\cite{lian2023explainable}. 
Video-LLaMA~\cite{zhang2023video} enables video-based comprehension by integrating audio-visual signals to LLaMA~\cite{touvron2023llama} and shows the ability to perceive and comprehend video content and generate meaningful responses. AffectGPT~\cite{lian2023explainable} pre-trains audio, video and text modalities together and handles the explainable multimodal emotion reasoning. 

The evolution of traditional multimodal sentiment analysis systems can be categorized by their approach to modality fusion.
The deep neural networks usually 
necessitate intricate fusion mechanisms to integrate complementary but also (possibly) redundant information across different modalities. 
Based on the summation of the work~\cite{zhu2023multimodal}, we introduce different fusion mechanisms. Early fusion, also named as feature-level fusion, extracts the various features from multiple modalities and fuses them at the input level. Late fusion, on the contrary, first conducts sentiment analysis based on each modality and then incorporates uni-modal sentiment decisions into the final prediction. 
Hybrid fusion combines early fusion and late fusion together. 
Tensor-based fusion obtains the multimodal sentence representation by tensor product of uni-modal representations. Translation-based fusion converts one modality into another to capture meaningful semantics across modalities. Memory-based fusion highlights the sequential interactions of each modality over time. 


Despite significant advances in multi-modal models, 
multi-modal emotion analysis encounters contradictions from single-modality emotion prediction.
This obstacle stems from the existence of potentially semantically conflicting information within each modality.
Human emotions are inherently expressed diversely across different modalities.
Occasionally, these modalities may convey incongruent meanings.
This phenomena complicates the integration and interaction of different modalities, inevitably posing difficulty for models. 
Consequently, the integration of these diverse modalities becomes a challenge and combating such inconsistency needs higher-level of multimodal reasoning. However, currently there lacks such a data-inconsistent benchmark dataset and no proper investigation into multi-modal emotion recognition models on handling inconsistent situations.
Our work fills the gap and contributes primarily on two key aspects: 


\begin{itemize}
\item We propose a clearly defined setting termed ``multi-modal conflicting data sentiment analysis" and introduce a standardized benchmark test set named DiffEmo for evaluation purposes with properly defined conflicted data. 
This initiative aims to establish a structured framework for researchers to assess and compare the performance of different models in handling modal conflicts effectively. 
\item We conduct a comprehensive evaluation of multiple models, including multi-modal large language models. Both deep neural networks with different modality fusion strategies and MLLMs fail to output reliable emotion label when facing inconsistent sentiment data. This extensive evaluation enables researchers to gain deeper insights into the strengths and limitations of various models when confronted with modal conflicts. Our results show that more complex reasoning ability is needed in multimodal sentiment analysis.  

\end{itemize}






\section{DiffEmo Dataset Construction} 
In multi-modal sentiment analysis, data conflict refers to the scenario in which a single data sample displays discordant emotional expressions across its three modalities: text, audio, and video. 
Various multi-modal sentiment datasets have been constructed to meet the growing demand for human sentiment perception~\cite{poria2018meld,zadeh2016multimodal,zadeh2018multimodal,liu2022make,lian2023mer}.
Among them, CH-SIMS v2.0~\cite{liu2022make} is a Chinese multi-modal sentiment analysis dataset. It contains a total of 4,402 video clips with both uni-modal and multi-modal annotations. Each instance in the dataset is annotated with sentiment scores ranging from \emph{Strong Negative (-1,-0.8)}, \emph{Weak Negative (-0.6,-0.4,-0.2)}, \emph{Neutral (0)}, \emph{Weak Positive (0.2, 0.4, 0.6)}, to \emph{Strong Positive (0.8, 1.0)} for uni-modal and multi-modal labels. 
From this dataset, we extract conflicting data samples based on their uni-modal annotations. A conflicting sample is defined as: 
\begin{align}
\text{Conflicting sample} = \left\{
\begin{array}{ll}
1, & \text{if } |m_1 - m_2| > 1 \\
0, & \text{otherwise}
\end{array}
\right.
\end{align}
where \( m_1 \) and \( m_2 \) represent the annotations from two different modalities. The absolute difference \( |m_1 - m_2| \) measures the inconsistency between these uni-modals. If the absolute difference exceeds 1, indicating strong inconsistency, the sample is labeled as conflicting (1); otherwise, it is labeled as consistent (0).

\begin{figure}[t]
  \centering
  \includegraphics[width=\linewidth]{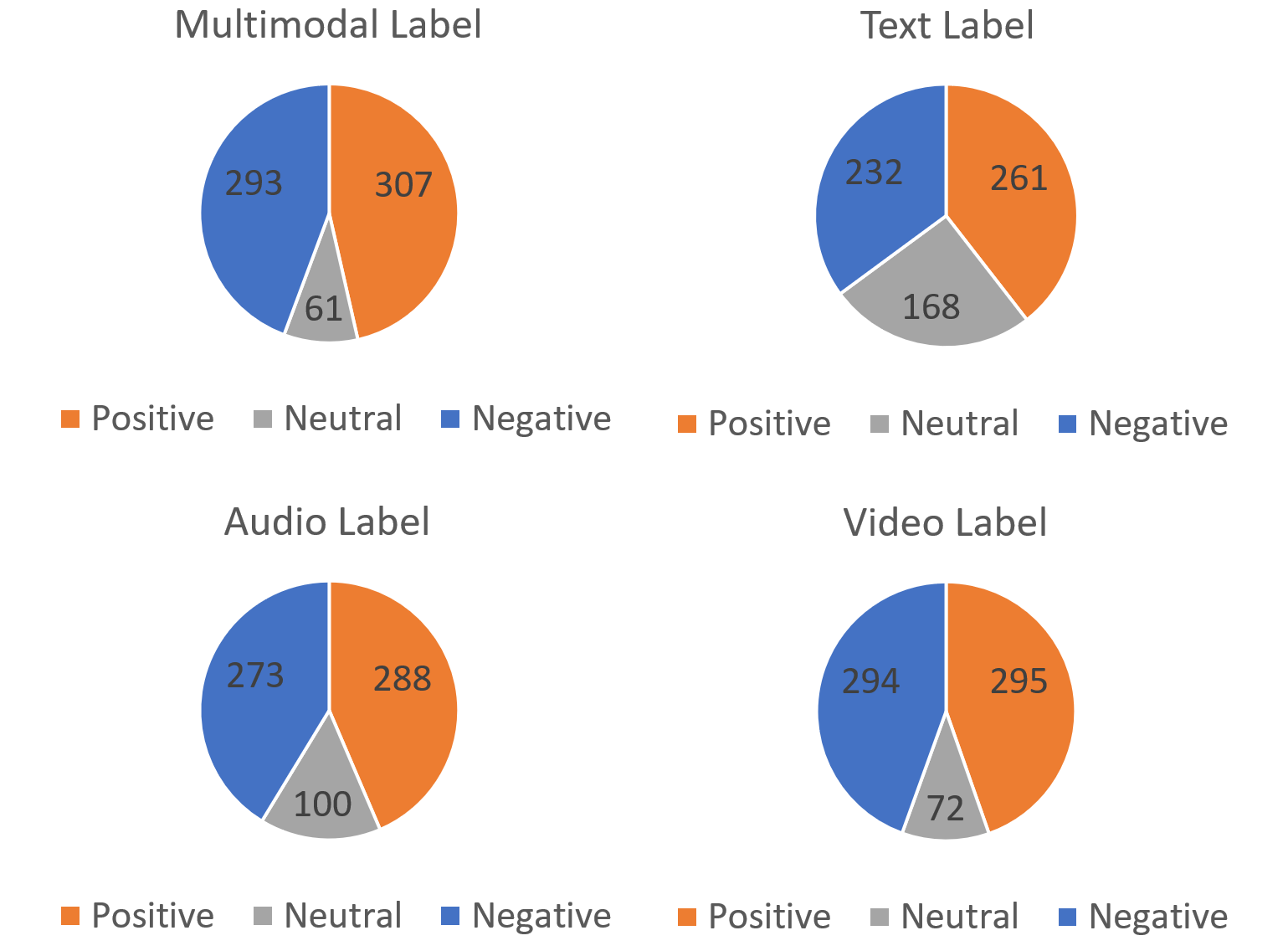}
  \caption{Distribution of uni-modal and multi-modal labels for conflicting data.} 
  \label{fig:label_count}
\end{figure}

Through this process, we collected a total of 661 conflicting data samples.
We present the label distribution of these samples in Figure~\ref{fig:label_count}. We find that text contains most neutral labels while video has higher distribution of positive sentiment. 
To further investigate the conflicting label distribution, we list the portion of each modality that is inconsistent with multi-modal labels when multi-modal label is classified as positive, neutral and negative in Table~\ref{tab:label_inconsistent}. We observe that textual modality is highly probably tends to be inconsistent with multi-modal label compared with other two modalities. 
Following manual scrutiny of the conflicting data, we identify and select a test set of 173 samples which encapsulates more severe semantic conflicts. 

\begin{table}[htbp]
\centering
\caption{
Illustration of modality inconsistency proportion, when a multi-modal label is categorized as positive, neutral, or negative, the portion of uni-modal label that contradicts the multi-modal label. 
}
\label{tab:label_inconsistent}
\begin{tabular}{|c|c|c|c|}
\hline
\multirow{2}{*}{Modality} & \multicolumn{3}{c|}{Multi-modal Label} \\ 
\cline{2-4} 
         & Positive & Neutral & Negative \\ \hline
T        & 25.87\%  & 8.78\% &  27.23\%    \\ \hline
A        & 14.52\%   & 7.56\% &  13.91\%   \\ \hline
V        & 16.79\%   & 8.01\% &  15.27\%   \\ \hline
\end{tabular}
\end{table}

In order to compare the performance of various models on conflicting data, we resample CH-SIMS v2.0 dataset and make a \textbf{DiffEmo Dataset}, consisting of three different test settings, namely Mixed Set, Conflicting Set and Aligned Set, aiming to validate that dealing with conflicting data indeed presents a more challenging setting. An illustration of data split for DiffEmo Dataset is shown in Table~\ref{tab:dataset}.

\begin{enumerate}
    \item Mixed Set, contains 2722 train samples, including 488 conflicting samples; 346 test samples, including 173 conflicting samples.
    \item Conflicting Set, contains the same train set as Mixed Set; 173 conflicting samples for test set. 
    \item Aligned Set, contains the same train set as Mixed Set; 173 coherent data samples for test set.
\end{enumerate}

\begin{table}[htbp]
\centering
\caption{Illustration of data split in DiffEmo Dataset.
}
\label{tab:dataset}
\begin{tabular}{|c|c|c|c|c|}
\hline
Set Name & \multicolumn{2}{c|}{Train } & \multicolumn{2}{c|}{Test} \\ 
\cline{2-5} 
         & Align & Conflict & Align & Conflict \\ \hline
Mixed     & 2234  & 488 &  173 & 173    \\ \hline
Conflicting         & 2234   & 488 &  0 & 173   \\ \hline
Aligned        & 2234   & 488 &  173 & 0   \\ \hline
\end{tabular}
\end{table}


\section{Evaluation Protocol}
In this section, we elaborate on the evaluation methodology utilized to address several inquiries concerning the performance of models on conflicting data. Initially, we validate that dealing with conflicting data indeed presents a more challenging setting compared to standard multi-modal emotion recognition. Subsequently, we conduct an analysis of various model performances and compare their fusion methods. Lastly, we assess the effectiveness of multi-modal large language models. Additionally, we perform an ablation study aimed at distinguishing between consistent data samples and conflicting data samples.


\noindent\textbf{Existence of conflicting data challenge}
By comparing the model performance between the Mixed Set, Conflicting Set and Aligned Set , we can demonstrate the existence of the conflicting data problem.


\noindent\textbf{Different fusion method comparison}
We conduct evaluations on multiple models to scrutinize the effectiveness of different fusion methods. 
A illustration is shown in Table~\ref{tab:fusion}.
We list the fusion method, whether it is multi-task, and performance of different models on original CH-SIMS v2.0 with evaluation metric of binary classification accuracy (Acc). They are mainstream multimodal emotion analysis models and exhibit good performance on the original dataset. 

\begin{table}[htbp]
\centering
\caption{An overview of the models for evaluation.}
\label{tab:fusion}
\begin{tabular}{|c|c|c|c|c|}
\hline
Model & \multicolumn{3}{c|}{Details} \\ 
\cline{2-4} 
                      & Fusion& Multi-task & Acc \\ \hline

EF-LSTM~\cite{williams2018a} & early &\textcolor{red}{\ding{55}} & 78.94\\ \hline 
SELF-MM~\cite{yu2021learning}       & early  & \textcolor{green}{\ding{51}}   &   79.69  \\ \hline
LF-DNN~\cite{cambria2018benchmarking}        &  late & \textcolor{red}{\ding{55}}  &  79.36   \\ \hline
MLF\_DNN~\cite{yu-etal-2020-ch} & late &\textcolor{green}{\ding{51}} & 76.59\\ \hline
MMIM~\cite{han2021improving}         & hybrid  & \textcolor{red}{\ding{55}}  &  76.6   \\ \hline
TETFN~\cite{wang2023tetfn}         & hybrid  & \textcolor{green}{\ding{51}}  &   79.73  \\ \hline
TFN~\cite{zadeh2017tensor}           & tensor  & \textcolor{red}{\ding{55}}   &  80.14   \\ \hline
MTFN~\cite{yu2020ch}         & tensor  & \textcolor{green}{\ding{51}}  &  80.43   \\ \hline

MLMF~\cite{liu2018efficient}          &  tensor & \textcolor{green}{\ding{51}}  &  77.23   \\ \hline
MulT~\cite{tsai2019multimodal}        & translation  & \textcolor{red}{\ding{55}}  &  80.68   \\ \hline
MFN~\cite{zadeh2018memory}          &  memory & \textcolor{red}{\ding{55}}  &  81.14   \\ \hline

Video-LLaMA~\cite{zhang2023video}  & \textcolor{red}{\ding{55}}  & \textcolor{green}{\ding{51}}  &  \textcolor{red}{\ding{55}}   \\ \hline
\end{tabular}
\end{table}

\noindent\textbf{Performance of multi-modal large language model}
We conduct an evaluation of the performance of multi-modal large language models on multi-modal conflicting data samples. This evaluation assesses the model's effectiveness in integrating and synthesizing information from various modalities within a specific context. Our objective is to test the model's capability to manage modal conflicts and generate coherent responses despite encountering conflicting semantics across different modalities.

\noindent\textbf{Ablation study: Distinguish conflicting data}
To explore the distinguish-ability between modal-conflicting data and modal-consistent data, we conduct an ablation study employing the Video-LLaMA for classification. This aims to ascertain whether instances of modal conflict exhibit discernible characteristics that set them apart from modal-consistent data.

\section{Experiments and Results}
In this part, we entail our experimental results for investigating traditional multimodal emotion models and LLMs under different data settings.
\subsection{Experimental details}
For traditional neural networks, we adopt consistent parameter settings with the approach outlined in prior work~\cite{mao2022m} and evaluate them on Mixed Set, Conflicting Set and Aligned Set. 
All experiments are conducted on one NVIDIA A10 GPU.

\begin{figure}[t]
  \centering
  \includegraphics[width=\linewidth]{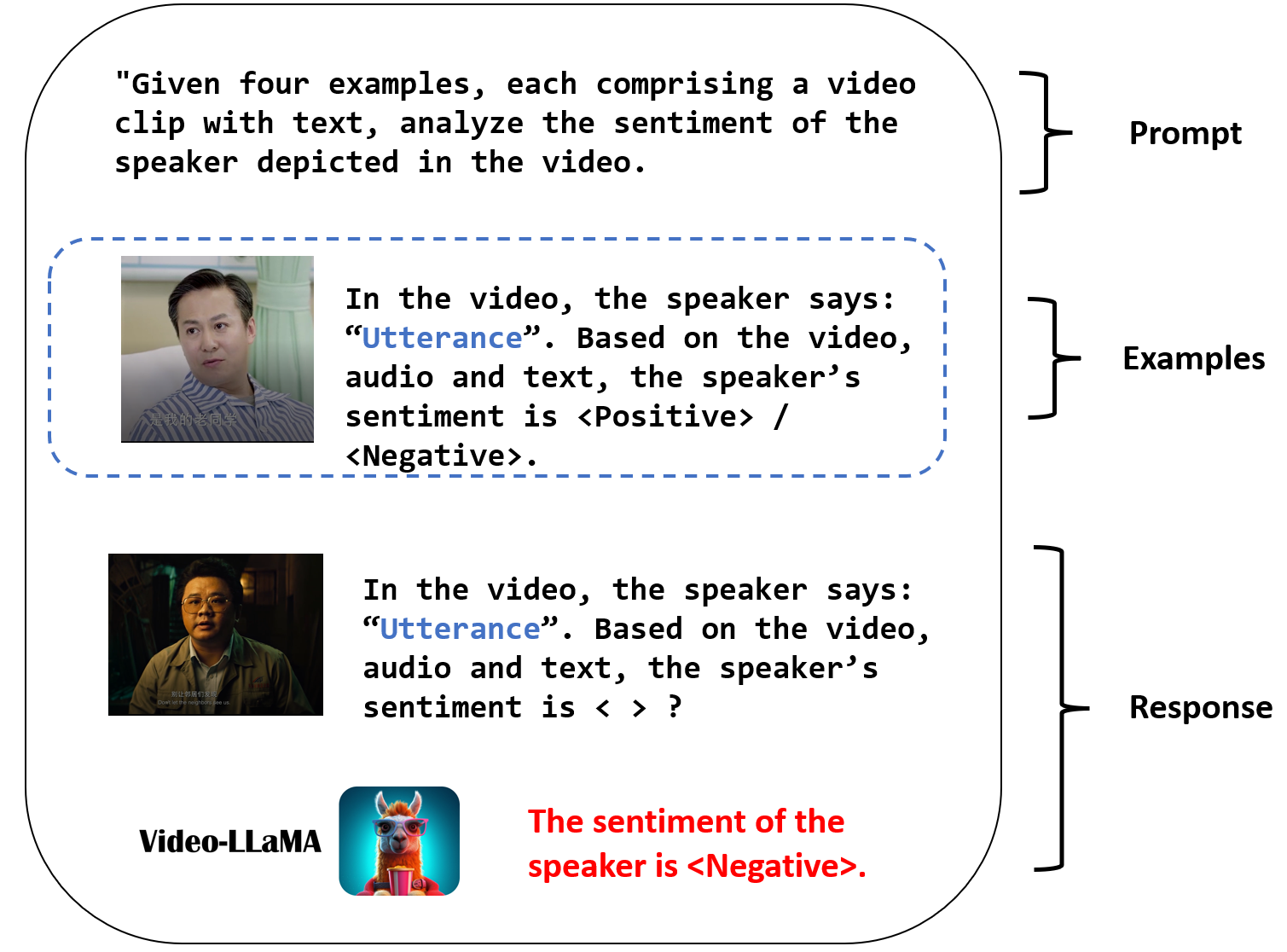}
  \caption{Prompt and in-context learning for Video-LLaMA.
  }
  \label{fig:mllm}
\end{figure}

Regarding Video-LLaMA, to leverage the knowledge of MLLM, we implement prompt and in-context learning strategies. Due to GPU memory limitations and the imperative need to incorporate acoustic, visual and textual  modalities, we choose Video-LLaMA-2-7B to test. To evaluate its emotion recognition ability, we employ prompt and in-context learning to unveil the latent knowledge embedded within the model. An illustration of the method is as shown in Figure~\ref{fig:mllm}. To facilitate comprehension of LLaMA, we use a translation tool~\cite{tang2020multilingual} to translate Chinese text to English, as the translation process in AffectGPT. Subsequently, we provide commands to prompt MLLM to follow the examples provided for in-context learning. These examples are randomly sampled from the dataset. We give it four examples for each test sample. Finally, we evaluate the model's ability on 200 randomly sampled test samples. 

\subsection{Evaluation metrics} 
We adopt the commonly-used metrics to evaluate the models. Binary classification accuracy (Acc)
reflects the
correctness of basic sentiment polarity prediction. Mean Absolute Error (MAE) is recorded for fine grained prediction evaluation.

\subsection{Results}

\begin{table*}[hbpt]
\centering
\caption{Performance comparison between different dataset settings. The numbers are shown as percentages.}
\label{tab:t1}

\begin{tabular}{ccccccc}
\hline
\multirow{2}{*}{\textbf{Model Name}} & \multicolumn{2}{c}{\textbf{Mixed Set}} & \multicolumn{2}{c}{\textbf{Conflicting Set}} & \multicolumn{2}{c}{\textbf{Aligned Set}}\\ 
\cline{2-7} 
& Acc & MAE & Acc 
  & MAE  & Acc 
  & MAE  \\ 
\hline

EF-LSTM &$77.48\pm1.43$ &$37.97\pm0.7$ & $67.62\pm2.34$ & $47.53\pm2.05$ &$85.9\pm1.99$ & $31.07\pm7.6$
\\ \hline
SELF-MM & $77.36\pm1.07$  & $37.84\pm1.01$ & $65.48\pm2.16$ & $49.36\pm2.08$ & $86.82\pm1.06$ & $26.93\pm0.57$ \\ 
\hline
LF-DNN & $75.07\pm3.27$ & $37.67\pm2.93$  & $62.98\pm4.94$ &  $50.39\pm3.46$ &$83.24\pm8.99$ &$29.62\pm7.24$  \\
\hline
MLF\_DNN &$75.78\pm3.68$ & $37.75\pm3.99$ &$67.38\pm2.3$ & $45.6\pm1.62$ & $84.63\pm8.24$  & $28.78\pm7.83$
\\ \hline
MMIM & $72.2\pm1.29$ & $42.06\pm0.52$ & $61.79\pm1.97$ & $49.99\pm1.72$ & $83.59\pm2.89$ & $32.0\pm1.01$ \\
\hline
TETFN & $76.42\pm0.8$  & $37.39\pm0.51$ & $66.07\pm1.36$  & $48.8\pm1.74$ & $89.37\pm0.28$  & $23.83\pm0.66$ \\ 
\hline
TFN & $67.98\pm7.47$ & $43.49\pm6.5$ & $64.52\pm3.72$  & $47.94\pm3.58$ & $75.15\pm8.72$ & $35.77\pm8.7$ \\
\hline
MTFN & $76.95\pm0.71$ & $35.84\pm0.95$ & $62.86\pm3.36$  & $49.32\pm2.11$ & $84.51\pm8.62$ & $28.37\pm7.86$  \\
\hline
MLMF & $69.91\pm5.05$ & $43.21\pm4.51$&$62.98\pm2.18$ &$49.52\pm1.81$ & $83.7\pm8.06$ &$30.39\pm7.38$ \\
\hline
MulT & $77.19\pm1.19$ &  $35.95\pm0.97$ & $65.6\pm1.48$  & $48.17\pm1.74$ & $89.13\pm1.18$ & $24.45\pm0.74$     \\ 
\hline
MFN & $73.55\pm5.23$ & $38.15\pm4.55$ & $66.07\pm1.06$ & $46.96\pm1.66$ &$81.39\pm8.08$  &$30.67\pm6.61$  \\
\hline
\end{tabular}
\end{table*}

\paragraph*{Performance comparison under different settings}
As shown in Table~\ref{tab:t1}, all models have a severe degradation compared between Conflicting Set and Aligned Set. We observe that for early and hybrid fusion, multi-task models always perform better compared with single-task ones, while for tensor-based and late fusion, single-task models achieve the best score, better than multi-task ones.
Testing on Conflicting Set, EF-LSTM and MLF\_DNN are best models, indicating that simple fusion networks sometimes may be effective for semantically non-aligned samples. 

As for the relatively degradation between three datasets, MulT performs relatively steady and achieves competitively performance on all three constructed dataset, which indicating the effectiveness of transformer architecture for capturing the intricate semantics for long-range series.

\paragraph*{Multimodal LLM on DiffEmo}

Due to the nature of LLM, they may occasionally produce irrelevant outputs, indicating a potential lack of contextual understanding.
We categorize the output of Video-LLaMA into three categories: ``True" if the output aligns with emotion label, ``False" if it deviates from the label, and ``None" if it is uncertain and gives irrelevant information.
We compare under different combination examples for in-context learning.
The results are shown in Table ~\ref{tab:t2}.

\begin{table}[hbpt]
\centering
\caption{Comparison between different in-context learning settings. The numbers are shown as percentages.}
\label{tab:t2}
\begin{tabular}{ccccc}
\hline
\multirow{2}{*}{\textbf{In-context}} & \multirow{2}{*}{\textbf{Test}} & \multicolumn{3}{c}{\textbf{Ratio}} \\ 
\cline{3-5} 
& & True & False & None  \\ 
\hline
Conflicting  & Conflicting & 43.4 & 39.8 &  16.6 \\ 
\hline
Conflicting  & Aligned & 43.3 & 37.5 & 19.2  \\ 
\hline
Aligned & Aligned & 52.1 & 26.7 & 21.2 \\ 
\hline
Mixed & Mixed & 43.8 & 37.9 & 18.3  \\ 
\hline

\end{tabular}
\end{table}

We observe that MLLM is struggling in emotion reasoning and fails to give the proper answer. 
When aligned data is given as the prompt, it also shows better results.
Due to complex human emotion expression, multimodal conflicting sentiment analysis is a harder setting for MLLM. Compared with deep neural networks, Video-LLaMA gets worse results. 
This may come from the insufficiency in video-based emotion entailment pre-training for MLLMs. 

\subsection{Ablation Study}
In order to distinguish the conflicting data with aligned ones, we test MLLM's ability to distinguish the inconsistency between various modalities.
Following previous settings, we change the prompt with ``Follow examples, decide whether the sentiment of the man in visual, acoustic and textual modalities shows inconsistent. Omit explanations". For in-context learning, we update the video and input that ``In the video, the speaker says: \textless utterance\textgreater. Based on the video, audio and speech of the speaker, the sentiment in visual, acoustic and textual modalities is \textless inconsistent\textgreater \ or \textless coherent\textgreater." The results is shown in Table ~\ref{tab:t3}. Video-LLaMA also encounters a great difficulty in distinguishing the conflicting data samples with those coherent ones. This indicates that conflicting data naturally shared a deceptive appearance with aligned ones. 
MLLM, interestingly, tends to recognize all samples as conflicting ones.
They lack a basic coherent recognition of the sentiment. Further training is needed on this aspect to enhance MLLMs' ability in emotion reasoning under conflicted conditions. 

\begin{table}[hbpt]
\centering
\caption{Results for distinguishing conflicting data with consistent data. The numbers are shown as percentages.}
\label{tab:t3}
\begin{tabular}{ccccc}
\hline
\multirow{2}{*}{\textbf{In-context}} & \multirow{2}{*}{\textbf{Test}} & \multicolumn{3}{c}{\textbf{Ratio}} \\ 
\cline{3-5} 
& & True & False & None  \\ 
\hline
mixed  & mixed & 53.68 & 42.10 & 3.16  \\ 
\hline

\end{tabular}
\end{table}

\section{Conclusion}
In this paper, driven by the ambiguity and subtlety in human emotions, we introduce a new setting called ``multimodal sentiment analysis with conflicting data'' and construct a well-split test set to explore the performance of traditional multimodal feature fusion based models and multimodal large language model. Compared with the performance of these models on normal setting, we observe a great degradation, which illustrates the difficulty of this challenge. 
We observe that simple fusion methods may be effective for conflicting data. Adding multi-setting learning to models may impact the performance depending on the fusion methods. Transformer is an effective method to capture semantics for conflicting data. For MLLMs, due to the complicated nature of human emotion reasoning, they suffer great degradation in detailing the emotion conveyed in videos. Distinguishing the semantically conflicting data samples with aligned ones may also be a challenge for MLLMs. In the future, video emotion entailment datasets can be constructed to enhance the emotion recognition ability for MLLMs.

\textbf{Limitations} Due to the GPU memory limitations and the necessity of using textual, acoustic and visual modalities, we only select Video-LLaMA to test for MLLMs, which may be insufficient. 

\bibliographystyle{IEEEtran}
\bibliography{mybib}

\begin{thebibliography}{10}
\providecommand{\url}[1]{#1}
\csname url@samestyle\endcsname
\providecommand{\newblock}{\relax}
\providecommand{\bibinfo}[2]{#2}
\providecommand{\BIBentrySTDinterwordspacing}{\spaceskip=0pt\relax}
\providecommand{\BIBentryALTinterwordstretchfactor}{4}
\providecommand{\BIBentryALTinterwordspacing}{\spaceskip=\fontdimen2\font plus
\BIBentryALTinterwordstretchfactor\fontdimen3\font minus \fontdimen4\font\relax}
\providecommand{\BIBforeignlanguage}[2]{{%
\expandafter\ifx\csname l@#1\endcsname\relax
\typeout{** WARNING: IEEEtran.bst: No hyphenation pattern has been}%
\typeout{** loaded for the language `#1'. Using the pattern for}%
\typeout{** the default language instead.}%
\else
\language=\csname l@#1\endcsname
\fi
#2}}
\providecommand{\BIBdecl}{\relax}
\BIBdecl

\bibitem{kaur2022multimodal}
R.~Kaur and S.~Kautish, ``Multimodal sentiment analysis: A survey and comparison,'' \emph{Research Anthology on Implementing Sentiment Analysis Across Multiple Disciplines}, pp. 1846--1870, 2022.

\bibitem{yang2020cm}
K.~Yang, H.~Xu, and K.~Gao, ``Cm-bert: Cross-modal bert for text-audio sentiment analysis,'' in \emph{Proceedings of the 28th ACM international conference on multimedia}, 2020, pp. 521--528.

\bibitem{li2022hybrid}
J.~Li \emph{et~al.}, ``Hybrid multimodal feature extraction, mining and fusion for sentiment analysis,'' in \emph{Proceedings of the 3rd International on Multimodal Sentiment Analysis Workshop and Challenge}, 2022, pp. 81--88.

\bibitem{franceschini2022multimodal}
R.~Franceschini \emph{et~al.}, ``Multimodal emotion recognition with modality-pairwise unsupervised contrastive loss,'' in \emph{2022 26th International Conference on Pattern Recognition (ICPR)}.\hskip 1em plus 0.5em minus 0.4em\relax IEEE, 2022, pp. 2589--2596.

\bibitem{ghosal2018contextual}
D.~Ghosal \emph{et~al.}, ``Contextual inter-modal attention for multi-modal sentiment analysis,'' in \emph{proceedings of the 2018 conference on empirical methods in natural language processing}, 2018, pp. 3454--3466.

\bibitem{wu2021text}
Y.~Wu \emph{et~al.}, ``A text-centered shared-private framework via cross-modal prediction for multimodal sentiment analysis,'' in \emph{Findings of the Association for Computational Linguistics: ACL-IJCNLP 2021}, 2021, pp. 4730--4738.

\bibitem{huang2021text}
P.~Huang \emph{et~al.}, ``Text sentiment analysis based on bert and convolutional neural networks,'' in \emph{Proceedings of the 2021 5th International Conference on Natural Language Processing and Information Retrieval}, 2021, pp. 1--7.

\bibitem{kenton2019bert}
J.~D. M.-W.~C. Kenton and L.~K. Toutanova, ``Bert: Pre-training of deep bidirectional transformers for language understanding,'' in \emph{Proceedings of naacL-HLT}, vol.~1, 2019, p.~2.

\bibitem{poklukar2022geometric}
P.~Poklukar \emph{et~al.}, ``Geometric multimodal contrastive representation learning,'' in \emph{International Conference on Machine Learning}.\hskip 1em plus 0.5em minus 0.4em\relax PMLR, 2022, pp. 17\,782--17\,800.

\bibitem{lv2021progressive}
F.~Lv \emph{et~al.}, ``Progressive modality reinforcement for human multimodal emotion recognition from unaligned multimodal sequences,'' in \emph{Proceedings of the IEEE/CVF Conference on Computer Vision and Pattern Recognition}, 2021, pp. 2554--2562.

\bibitem{sun2023layer}
J.~Sun \emph{et~al.}, ``Layer-wise fusion with modality independence modeling for multi-modal emotion recognition,'' in \emph{Proceedings of the 61st Annual Meeting of the Association for Computational Linguistics (Volume 1: Long Papers)}, 2023, pp. 658--670.

\bibitem{lian2024merbench}
Z.~Lian \emph{et~al.}, ``Merbench: A unified evaluation benchmark for multimodal emotion recognition,'' \emph{arXiv preprint arXiv:2401.03429}, 2024.

\bibitem{lian2023explainable}
Z.~Lian, L.~Sun, M.~Xu, H.~Sun, K.~Xu, Z.~Wen, S.~Chen, B.~Liu, and J.~Tao, ``Explainable multimodal emotion reasoning,'' \emph{arXiv preprint arXiv:2306.15401}, 2023.

\bibitem{zhang2023video}
H.~Zhang, X.~Li, and L.~Bing, ``Video-llama: An instruction-tuned audio-visual language model for video understanding,'' \emph{arXiv preprint arXiv:2306.02858}, 2023.

\bibitem{touvron2023llama}
H.~Touvron \emph{et~al.}, ``Llama: Open and efficient foundation language models,'' \emph{arXiv preprint arXiv:2302.13971}, 2023.

\bibitem{zhu2023multimodal}
L.~Zhu \emph{et~al.}, ``Multimodal sentiment analysis based on fusion methods: A survey,'' \emph{Information Fusion}, vol.~95, pp. 306--325, 2023.

\bibitem{poria2018meld}
S.~Poria \emph{et~al.}, ``Meld: A multimodal multi-party dataset for emotion recognition in conversations,'' \emph{arXiv preprint arXiv:1810.02508}, 2018.

\bibitem{zadeh2016multimodal}
A.~Zadeh \emph{et~al.}, ``Multimodal sentiment intensity analysis in videos: Facial gestures and verbal messages,'' \emph{IEEE Intelligent Systems}, vol.~31, no.~6, pp. 82--88, 2016.

\bibitem{zadeh2018multimodal}
A.~B. Zadeh \emph{et~al.}, ``Multimodal language analysis in the wild: Cmu-mosei dataset and interpretable dynamic fusion graph,'' in \emph{Proceedings of the 56th Annual Meeting of the Association for Computational Linguistics (Volume 1: Long Papers)}, 2018, pp. 2236--2246.

\bibitem{liu2022make}
Y.~Liu \emph{et~al.}, ``Make acoustic and visual cues matter: Ch-sims v2. 0 dataset and av-mixup consistent module,'' in \emph{Proceedings of the 2022 International Conference on Multimodal Interaction}, 2022, pp. 247--258.

\bibitem{lian2023mer}
Z.~Lian \emph{et~al.}, ``Mer 2023: Multi-label learning, modality robustness, and semi-supervised learning,'' in \emph{Proceedings of the 31st ACM International Conference on Multimedia}, 2023, pp. 9610--9614.

\bibitem{williams2018a}
J.~Williams \emph{et~al.}, ``Recognizing emotions in video using multimodal dnn feature fusion,'' in \emph{Proceedings of Grand Challenge and Workshop on Human Multimodal Language (Challenge-HML)}.\hskip 1em plus 0.5em minus 0.4em\relax Association for Computational Linguistics, 2018, pp. 11--19.

\bibitem{yu2021learning}
W.~Yu, H.~Xu, Z.~Yuan, and J.~Wu, ``Learning modality-specific representations with self-supervised multi-task learning for multimodal sentiment analysis,'' in \emph{Proceedings of the AAAI conference on artificial intelligence}, vol.~35, no.~12, 2021, pp. 10\,790--10\,797.

\bibitem{cambria2018benchmarking}
E.~Cambria \emph{et~al.}, ``Benchmarking multimodal sentiment analysis,'' in \emph{Computational Linguistics and Intelligent Text Processing: 18th International Conference, CICLing 2017, Budapest, Hungary, April 17--23, 2017, Revised Selected Papers, Part II 18}.\hskip 1em plus 0.5em minus 0.4em\relax Springer, 2018, pp. 166--179.

\bibitem{yu-etal-2020-ch}
\BIBentryALTinterwordspacing
W.~Yu, H.~Xu, F.~Meng, Y.~Zhu, Y.~Ma, J.~Wu, J.~Zou, and K.~Yang, ``{CH}-{SIMS}: A {C}hinese multimodal sentiment analysis dataset with fine-grained annotation of modality,'' in \emph{Proceedings of the 58th Annual Meeting of the Association for Computational Linguistics}, D.~Jurafsky, J.~Chai, N.~Schluter, and J.~Tetreault, Eds.\hskip 1em plus 0.5em minus 0.4em\relax Online: Association for Computational Linguistics, Jul. 2020, pp. 3718--3727. [Online]. Available: \url{https://aclanthology.org/2020.acl-main.343}
\BIBentrySTDinterwordspacing

\bibitem{han2021improving}
W.~Han, H.~Chen, and S.~Poria, ``Improving multimodal fusion with hierarchical mutual information maximization for multimodal sentiment analysis,'' in \emph{Proceedings of the 2021 Conference on Empirical Methods in Natural Language Processing}, 2021, pp. 9180--9192.

\bibitem{wang2023tetfn}
D.~Wang \emph{et~al.}, ``Tetfn: A text enhanced transformer fusion network for multimodal sentiment analysis,'' \emph{Pattern Recognition}, vol. 136, p. 109259, 2023.

\bibitem{zadeh2017tensor}
A.~Zadeh \emph{et~al.}, ``Tensor fusion network for multimodal sentiment analysis,'' \emph{arXiv preprint arXiv:1707.07250}, 2017.

\bibitem{yu2020ch}
W.~Yu \emph{et~al.}, ``Ch-sims: A chinese multimodal sentiment analysis dataset with fine-grained annotation of modality,'' in \emph{Proceedings of the 58th annual meeting of the association for computational linguistics}, 2020, pp. 3718--3727.

\bibitem{liu2018efficient}
Z.~Liu \emph{et~al.}, ``Efficient low-rank multimodal fusion with modality-specific factors,'' \emph{arXiv preprint arXiv:1806.00064}, 2018.

\bibitem{tsai2019multimodal}
Y.-H.~H. Tsai \emph{et~al.}, ``Multimodal transformer for unaligned multimodal language sequences,'' in \emph{Proceedings of the conference. Association for Computational Linguistics. Meeting}, vol. 2019.\hskip 1em plus 0.5em minus 0.4em\relax NIH Public Access, 2019, p. 6558.

\bibitem{zadeh2018memory}
A.~Zadeh \emph{et~al.}, ``Memory fusion network for multi-view sequential learning,'' in \emph{Proceedings of the AAAI conference on artificial intelligence}, vol.~32, no.~1, 2018.

\bibitem{mao2022m}
H.~Mao \emph{et~al.}, ``M-sena: An integrated platform for multimodal sentiment analysis,'' \emph{arXiv preprint arXiv:2203.12441}, 2022.

\bibitem{tang2020multilingual}
Y.~Tang \emph{et~al.}, ``Multilingual translation with extensible multilingual pretraining and finetuning,'' \emph{arXiv preprint arXiv:2008.00401}, 2020.

\end{thebibliography}

\end{document}